\documentclass[letterpaper, 10 pt, journal, twoside]{IEEEtran}
\ifCLASSINFOpdf
\else
\fi
\usepackage{amsmath}
\usepackage{amsfonts}
\usepackage{mathtools}

\usepackage{ marvosym } 
\usepackage{xcolor}
\usepackage{hyperref}
\usepackage[numbers]{natbib}
\usepackage{graphicx}
\usepackage{float} 
\usepackage{algorithm}
\usepackage{algpseudocode}
\usepackage[nice]{nicefrac}
\usepackage{multirow}  

\def\ie{\emph{i.e.}}
\def\eg{\emph{e.g.}}

\def\RR{\mathbb{R}}

\def\gam{{\gamma}}

\def\hgam{{\hat{\gam}}}

\def\cM{{\mathcal{M}}}

\def\cS{{\mathcal{S}}}

\def\cE{{\mathcal{E}}}
\def\cD{{\mathcal{D}}}
\def\Serp{{\cS_{\rm serp}}}

\DeclareMathOperator{\SE}{SE}
\DeclareMathOperator{\SO}{SO}

\newcommand{\figref}[1]{Fig.~\ref{#1}}

\newcommand{\teqref}[1]{Eq.~(\ref{#1})}

\newcommand{\teqrefss}[3]{Eqs.~(\ref{#1}, \ref{#2}, \ref{#3})}
\newcommand{\secref}[1]{Sec.~\ref{#1}}

\renewcommand{\algref}[1]{Alg.~\ref{#1}}

\usepackage{algorithm}
\usepackage{algorithmicx}
\usepackage{algpseudocode}

\algnewcommand\algorithmicinput{\textbf{Input:}}
\algnewcommand\algorithmicoutput{\textbf{Output:}}
\algnewcommand\Input{\item[\algorithmicinput]}%
\algnewcommand\Output{\item[\algorithmicoutput]}%

\usepackage[normalem]{ulem}
\usepackage{soul}

\newcommand{\PM}[1]{{\textcolor[rgb]{1.0,0.53,0.0}{(\textbf{PM:} #1)}}}
\newcommand{\OG}[1]{{\textcolor[rgb]{1,0.0,0.5}{(\textbf{OG:} #1)}}}

\begin{document}
%
\title{A variational approach to geometric mechanics\\ for undulating robotic locomotion}

%
%

\author{Sean~Even\(^1\), 
        Patrick~S.~Martinez\(^2\), 
        Cora~Keogh\(^3\), 
        Oliver~Gross\(^{4,*}\), 
        Yasemin~\"Ozkan-Ayd\i n\(^{1,*}\) 
        and Peter~Schr\"oder\(^2\)
\thanks{\(^1\)University of Notre Dame, \(^2\)California Institute of Technology, \(^3\)Dublin City University, \(^4\)EPFL (Swiss Federal Institute of Technology, Lausanne)\\ \(^*\)Corresponding authors. }%

}

\maketitle

\begin{abstract} 
Limbless organisms of all sizes use undulating patterns of self-deformation to locomote. Geometric mechanics, which maps deformations to motions, provides a powerful framework to formalize and investigate the theoretical properties and limitations of such modes of locomotion. However, the inherent level of abstraction poses a challenge when bridging the gap between theory or simulations and laboratory experiments. 
We investigate the challenges of modeling motion trajectories of an undulating robotic locomotor by comparing experiments and simulations performed with a variational integrator. Despite the extensive simplifications that the model based on a geometric variation principle entails, the simulations show good agreement on average. Notably, our approach merely requires the knowledge of the \emph{dissipation metric}---a Riemannian metric on the configuration space, which can in practice be approximated by means closely resembling \emph{resistive force theory}.
\end{abstract}


\IEEEpeerreviewmaketitle

\section{Introduction}
\IEEEPARstart{U}{}nderstanding how animals navigate and propel themselves through diverse environments is a fundamental question in biomechanics \cite{dickinson2000animals}. The dynamics of locomotion are intertwined with environmental factors---complex interactions between an organism’s morphology and the surrounding medium critically influence performance metrics such as velocity, maneuverability, and energy expenditure. By studying these interactions, we gain valuable insights into how animals have evolved to optimize their movement, which can, in turn, inspire innovative designs in robotics and other fields seeking to mimic or leverage these natural strategies.

Among all organisms, snakes stand out for their exceptional versatility in locomotion. Navigating diverse environments—from dense underbrush and rocky crevices to smooth, sandy surfaces---with remarkable adaptability~\cite{Gray:1953:UP}. They employ various locomotion strategies, including undulating, sidewinding, and concertina movements~\cite{Gray:1946:MLS, Jayne:1986:KTSL} with each gait finely tuned to overcome specific environmental challenges. This adaptability not only allows them to efficiently exploit a wide range of ecological niches but also highlights the sophistication of their biomechanical systems, which contribute to their widespread success as a species across varied habitats~\cite{gans1962terrestrial,Edwards1985,Gans1986,lillywhite2014snakes}. Their unique locomotion strategies have inspired numerous robotic studies, aiming to replicate these natural mechanisms for advanced robotic mobility\cite{Hirose2009,Shao2015,walker2016snake,pettersen2017snake,liu2021review,yang2022snake}.

Robotic snake locomotion has been heavily influenced by the application of geometric mechanics, which provides a unified framework for controlling these dynamic systems~\cite{Shapere:1989:GKD, Ostrowski:1998:GMU, Gir+:2013:COS, Alouges:2019:EOS, Chong:2023:OCP, Rieser:2024:GPp}. One of the foundational models, proposed by \citet{Hirose:1993:SSS}, utilizes a serpenoid shape to represent the continuous curve of a snake's body, allowing for precise control of its undulatory motion. This geometric approach has been extended by \citet{Branyan:2017:SSR}, who implemented a gait optimization strategy to enhance locomotion efficiency, through coordinated shape changes. Similarly, \cite{Ramasamy:2019:GOG} applied soap--bubble optimization to identify optimal gaits based on the curvature of the shape space, while~\cite{Hatton:2017:KCE} optimized gaits for their power consumption based on a \emph{power dissipation metric}. Expanding these principles to more complex terrains, \citet{Dai:2016:GSGS} optimized robotic snake gaits on granular surfaces, leveraging a geometric framework to simplify the dynamic interaction between the robot and the shifting terrain. These control approaches demonstrate the flexibility and effectiveness of geometric mechanics in enabling precise, efficient locomotion across various environments.


The inherent level of abstraction, however, can pose a challenge when geometric mechanics-based simulations are to be adapted to reproduce or predict real-world experiments. 
\begin{figure}[t!]
    \centering
    \includegraphics[width = \columnwidth]{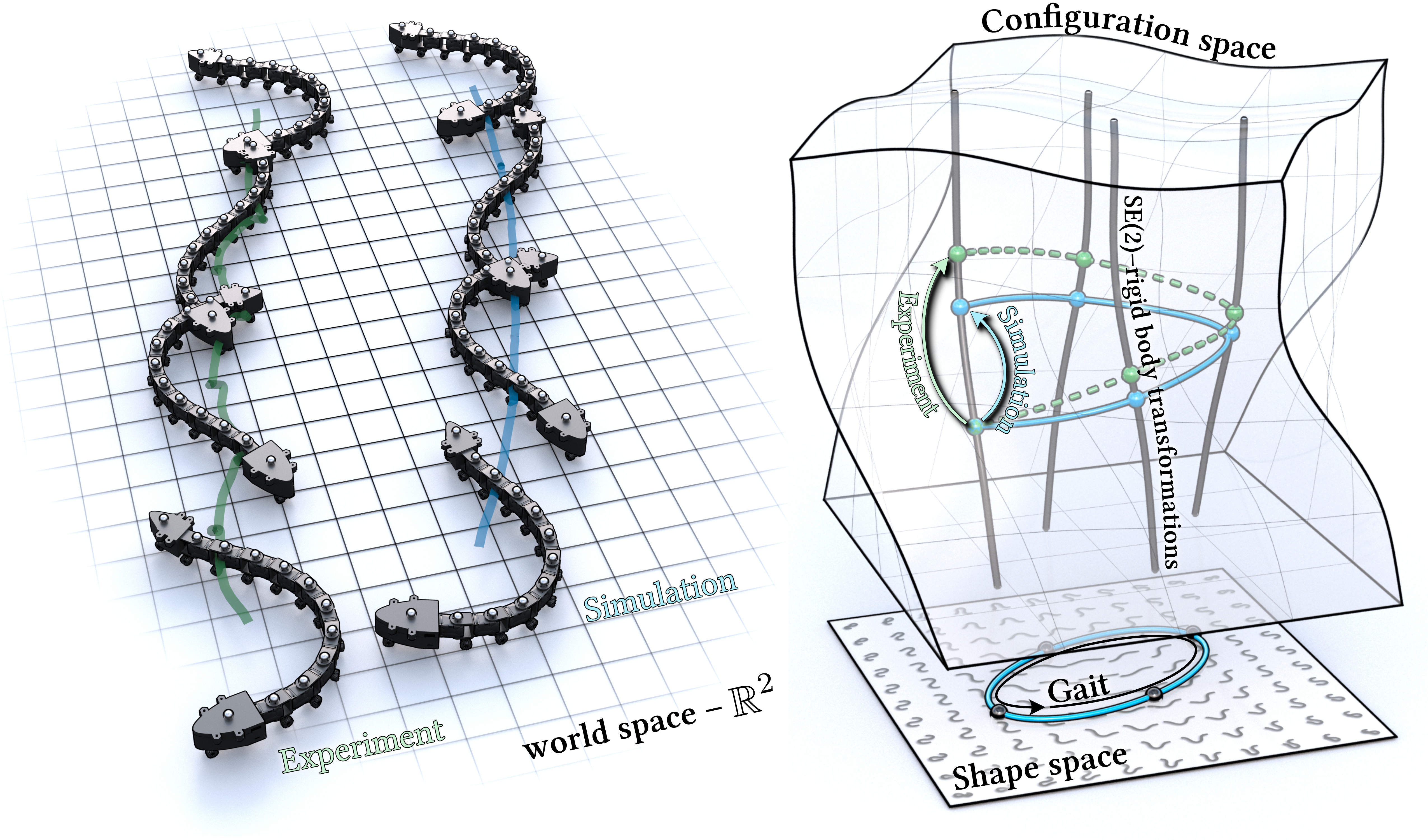}
    \vspace{-7mm}
    \caption{Geometric mechanics maps gaits of undulatory robotic locomotors to motion trajectories \emph{(blue)} in world coordinates. The accuracy of this map compared to laboratory experiments \emph{(green)} depends on the choice of model parameters, for which the Riemannian metric on the configuration space provides a natural description.} 
        \vspace{-5mm}
    \label{fig:HeroShot}
\end{figure}
This paper addresses the challenges of bridging the gap between abstract theoretical models and empirical validation using a practical example. Restricting our attention to the case of undulating locomotion, we 
\begin{itemize}
    \item calibrate a general simulation framework~\cite{Gross:2023:MFSC} to match motion trajectories of a snake-like robot.
    \item employ an approximation of the dissipation metric which closely resembles \emph{Resistive Force Theory} (RFT).  
    \item employ a variational integrator, derived from a geometric variational principle, to integrate motion trajectories from given shape sequences. 
    \item compare the locomotive performance of simulations and laboratory experiments for various gaits. In particular, we validate if performance differences of the gaits carry over to the real world.
\end{itemize}
This allows us to test whether a highly simplified model can still accurately describe the system's behavior, aiming to replace costly detailed simulations with more efficient calculations. 

A convenient calibration of the geometric mechanics model for a specific laboratory setup enables a more efficient approximation of the trajectories. In particular, by demonstrating the effectiveness of an intuitive approach to geometric mechanics that only requires the specification of a Riemannian metric in the form of RFT-like calibrations, we show the practical accessibility to the multitude of benefits this abstract setup provides. Ultimately, this paves the way to new areas of application such as inverse design or reinforcement learning~\cite{Bing:2022:STR}. 

\section{Preliminaries}
Geometric mechanics analyzes dynamical systems based on a special structure of the configuration space \(\cM\). For any state of a locomotive system in \(\RR^3\) the locomotors \emph{shape} \(\hgam\in \cS\) is treated distinguished from its respective \emph{positioning}  \(\gam\in\cM\). Two positioned shapes have the same shape if they differ only by a rigid body motion \(g\in \SE(3)\) in world space\footnote{Although the locomotion that we consider in the present paper is limited to two dimensions, we treat it as a special case of the three-dimensional case, which allows us to establish a more general set of equations that was used for the numerical implementation.}. That is, \(g\colon\RR^3\to\RR^3,\, x\mapsto Ax+b\) 
for some \(A\in\SO(3)\) and \(b\in\RR^3\).

As a result, the configuration space \(\cM\) decomposes into six-dimensional \emph{fibers}, 
\begin{equation}
    \label{eq:FiberDef}
    \{\gamma=g(\hgam)\mid g\in\SE(3)\}\subset\cM,
\end{equation}
consisting of all possible positions of a body posture \(\hgam\in\cS\). 

Therefore, the dynamics of an undulating robotic locomotor can be described by a smooth curve \(\hgam\colon [0,T]\to \cS\) into the \emph{shape space} \(\cS \) 
together with a smooth map \(g\colon [0,T]\to \SE(3)\) positioning each shape in world space. A path \(\gam\colon [0,T]\to\cM\) in the configuration space is said to be a \emph{lift} of \(\hgam\) if \(\pi(\gam)=\hgam\), where \(\pi\colon\cM\to\cS\) denotes the projection from the configuration space onto the shape space. 
By decoupling shapes from their position, optimization and control problems in robotics can often be reduced to a problem on the shape space \(\cS\). 
 

\subsection{Geometric locomotion}
For scenarios involving, \eg, highly damped environments or granular media one can exploit a linear relationship between a system's shape changes \(\hat\gamma'\) and its position to integrate the motion trajectory of a shape changing body~\cite{Ostrowski:1998:GMU, Ramasamy:2019:GOG}. Typically, a \emph{local connection form} \(\varpi(\hat\gamma)\) 
is used to map infinitesimal shape changes to body velocities in world coordinates by 
\begin{align}
    \label{eq:ReconstructionEquation}
    \gamma' = -\varpi(\hat\gamma)\,\hat\gamma'.
\end{align}
\teqref{eq:ReconstructionEquation} assures that any \(1\)-parameter family \(t\mapsto \hgam_t\in\cS\) of body poses can, up to a global rigid body motion, be uniquely lifted to a \(1\)-parameter family \(t\mapsto\gam_t\in\cM\) describing the motion trajectory of the shape changing body.

When we consider a closed loop in the shape space \(\cS\), it describes a periodic sequence of shapes, which is referred to as a \emph{gait}. Despite the periodic nature of gaits, the resulting lift \(\gam\) does in general not close up  (\figref{fig:HeroShot}). This aperiodicity is precisely why the geometric description of the dynamical system can lead to a net displacement of shape-changing bodies. The resulting net displacement of the locomotor after one gait cycle is given by \(g(0)^{-1}g(T)\in\SE(n)\) (\figref{fig:HeroShot}) and the extent of this displacement depends on the geometry of the fiber bundle~\cite{Frankel:2011:GOP}.

\subsection{A variational approach}
\label{sec:VariationalIntegrator}
Instead of numerically integrating \teqref{eq:ReconstructionEquation}, \emph{variational integrators} provide an alternative approach to integrating motion trajectories from shape sequences. They are derived following the guiding principle to \emph{first discretize and then optimize}~\cite{Marsden:2001:DMV}. By construction, they exhibit several advantageous properties as they are automatically symplectic and momentum preserving, and exhibit good energy behavior for exponentially long times~\cite{Marsden:2001:DMV, Leok:2015:VI}.

A unified framework that makes variational integrators straightforward to use for a variety of scenarios---including the one we examine---has been proposed by~\citet{Gross:2023:MFSC}. In this section, we will briefly outline the key aspects of this approach, while for a more detailed exposition and derivations, we refer to the original reference. Notably, this approach does not require explicit prior knowledge of the local connection forms. 
Instead, it merely requires knowledge of the dissipation metric on \(\cM\) for which even a rough approximation in the spirit of RFT~\cite{Gray:1955:PSU, Zhang:2014:TEO} has proven sufficient in practice.

\subsubsection{Discretization and local dissipation metric}
\begin{figure}
    \centering
    \includegraphics[width=\linewidth]{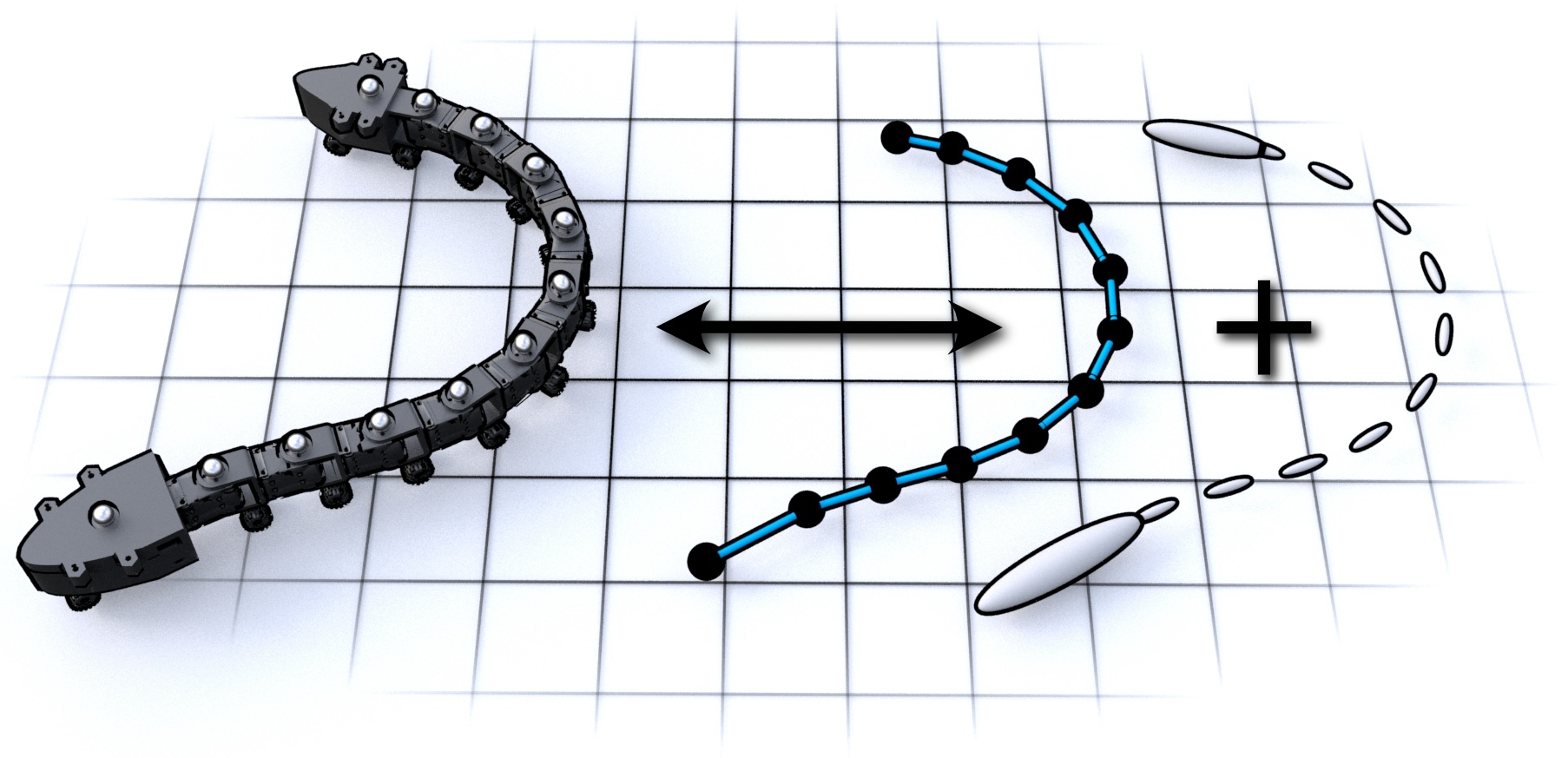}
        \vspace{-7mm}
    \caption{We represent the robotic system comprised of linked elements as a polygonal curve. Displacements of vertices anisotropically dissipate energy to the environment which is represented by local dissipation tensors (see \secref{sec:VariationalIntegrator})}
        \vspace{-5mm} \label{fig:LocalDissipationTensor}
\end{figure}

We discretize the snake-like undulating locomotor at a given time \(t\) as a polygonal curve \(\gam^t\) consisting of \(N\) vertices in \(\RR^3\) (\figref{fig:LocalDissipationTensor}). Moreover, we restrict our attention to time-discrete sequences of discrete positioned shapes, \ie, an indexed sequence \(\gam^0,\ldots,\gam^T\in \cM=(\RR^3)^N.\) 
To each vertex \(\gam^t_k\) of a curve \(\gam^t\) we assign a unit tangent vector \(T^t_k\in S^2\). When a positioned shape is transformed by a rigid motion \(\gam\mapsto A\gam + b\), the tangent vectors at the vertices are transformed as well by \(T\mapsto AT\).

Displacing a body in a viscous medium causes energy dissipation due to, \eg, viscous drag. In the spirit of RFT, we approximate this energy dissipation by summing up the energy dissipated from displacing individual vertices, thus neglecting any effects from interactions. Treating vertices as \emph{local dissipation elements} (\figref{fig:LocalDissipationTensor}) we associate them with anisotropic \emph{local dissipation tensors}, \ie, symmetric and positive definite blocks of the form 
\begin{align}
    \label{eq:LocalDissipationTensor}
    D_k^t\coloneqq w_k(I + (\epsilon - 1) T^t_k\otimes T^t_k) \in\RR^{3,3}. 
\end{align}
Here, \(w_k\) are \emph{integration weights}, while the \emph{anisotropy ratio} \(\epsilon\in(0,1]\) controls the ease of tangential motion compared to normal motion (\figref{fig:LocalDissipationTensor}). For \(\epsilon\approx0\) tangential displacements cause close to no energy dissipation, while for \(\epsilon=1\) the tensor becomes isotropic and there are no preferred directions and no net displacement can be achieved~\cite{Gray:1955:PSU}.

\subsubsection{Variational energy}
The discrete variational energy we consider is of the form
\begin{align}
    \label{eq:VariationalEnergy}
    \cE(\gam^0,\ldots,\gam^T)=\tfrac{1}{2}\sum_{t=0}^{T-1} \langle \cD^{(t,t+1)} \Delta p^{(t,t+1)}, \Delta p^{(t,t+1)} \rangle.
\end{align}
Here, denoting the concatenation of the vertex positions of \(\gam^t\) by \(p^t\), \(\Delta p^{(t,t+1)} \coloneqq p^{t+1}-p^t\) and \(\cD^{(t,t+1)}\) is symmetric, positive definite and models the dissipation metric. We follow the definition in \cite{Gross:2023:MFSC} and choose \(\cD^{(t,t+1)}\coloneqq \tfrac{1}{2}(\cD^t + \cD^{t+1})\) where \(\cD^t\in\RR^{3N,3N}\) are block-diagonal matrices with blocks of the form given in \teqref{eq:LocalDissipationTensor}. This energy measures the \emph{total energy dissipation} affected by the displacements \(\Delta p^{(t,t+1)}\). 

\subsubsection{Variational integrator}
A physical motion extremizes \teqref{eq:VariationalEnergy} under variations by rigid body transformations~\cite{Gross:2023:MFSC}. The resulting Euler-Lagrange equation corresponds to the non-holonomic constraints of the system~\cite{Marsden:2001:DMV}. For bodies initially at rest, they are given by 
\begin{equation}
    \label{eq:NonHolonomicConstraintsMu}
\mu(\gam^t,\gam^{t+1})=0.
\end{equation}
Here, \(\mu\) is the \emph{geometric momentum}, which is defined as 
\begin{align}
    \label{eq:ELEq}
    \mu(\gam^t,\gam^{t+1}) \coloneqq -\tfrac{1}{2}\!\sum_{k}
    \begin{pmatrix}
    \mu_{\rm rot}(\gam^t,\gam^{t+1})_k \\
    \mu_{\rm tran}(\gam^t,\gam^{t+1})_k
  \end{pmatrix},
\end{align}
where 
\begin{align*}
    \mu_{\rm rot}(\gam^t,\gam^{t+1})_k\coloneqq & \ p^{t+1}_k\times(D^{t}\Delta
    p^{(t,t+1)})_k \\ &+p^{t}_k\times(D^{t+1}\Delta p^{(t,t+1)})_k\\
    \mu_{\rm tran}(\gam^t,\gam^{t+1})_k \coloneqq &\ (D^{(t,t+1)}\Delta p^{(t,t+1)})_k.
\end{align*}

Since the shapes are considered to be given, for every timestep, the six constraints in \teqref{eq:NonHolonomicConstraintsMu} match the six degrees of freedom for their respective position (\teqref{eq:FiberDef}). Therefore,  \teqref{eq:NonHolonomicConstraintsMu} can be solved numerically exactly at each time step (\algref{alg:IntegrateMotion}), thus avoiding the accumulation of integration errors. In practice, given two consecutive shapes together with their respective material parameters, \teqref{eq:NonHolonomicConstraintsMu} is satisfied for roots of 
\begin{equation}
    \label{eq:RootFindingProblem}
    \SE(3)\to\RR^{6},\ g_t\mapsto \mu(\gam^t,g_t(\hgam^{t+1})).
\end{equation}
That is, \(\hgam_{t+1}\) is positioned by \(g_t\in \SE(3)\) such that \teqref{eq:NonHolonomicConstraintsMu} is satisfied. 
\begin{algorithm}
\caption{--- \textbf{IntegrateMotionTrajectory} 
\cite{Gross:2023:MFSC}
}
\label{alg:IntegrateMotion}
\begin{algorithmic}[1]
\Input shapes \((\hgam^{0},\ldots,\hgam^{T})\), parameters \((w_0,\ldots, w_N, \epsilon)\)
\Output positioned shapes \(\gam^{0},\ldots,\gam^{T}\)
\State \(\gam^0\gets \hgam^0\)
\For{\(t=1,\ldots,T\)}
\State \(g_{t}\gets\) solve \(\mu(\gam^{t-1},g_t(\hgam^t)) =0\) \Comment \teqrefss{eq:NonHolonomicConstraintsMu}{eq:ELEq}{eq:RootFindingProblem} \label{step:SolveForRoot}
\State \(\gam^{t}\gets g_t(\hgam^t)\)
\EndFor
\end{algorithmic}
\end{algorithm}
For three spacial dimensions, determining such a positioning amounts to solving a non-linear system in six variables (\teqref{eq:RootFindingProblem}), while in two spacial dimensions only three variables are left. 




\section{Shape space and gait design}
We describe the examined gaits with help of the space \(\Serp\) of \emph{serpenoid} curves, which is commonly used to describe undulations of non-anthropomorphic organisms in low-Reynolds number environments~\cite{Hirose:1993:SSS, Rieser:2024:GPp} and highly damped environments \cite{Rieser:2024:GPp, chong2022coordinating}. 

\begin{figure}[h]
    \centering
\includegraphics[width=\columnwidth]{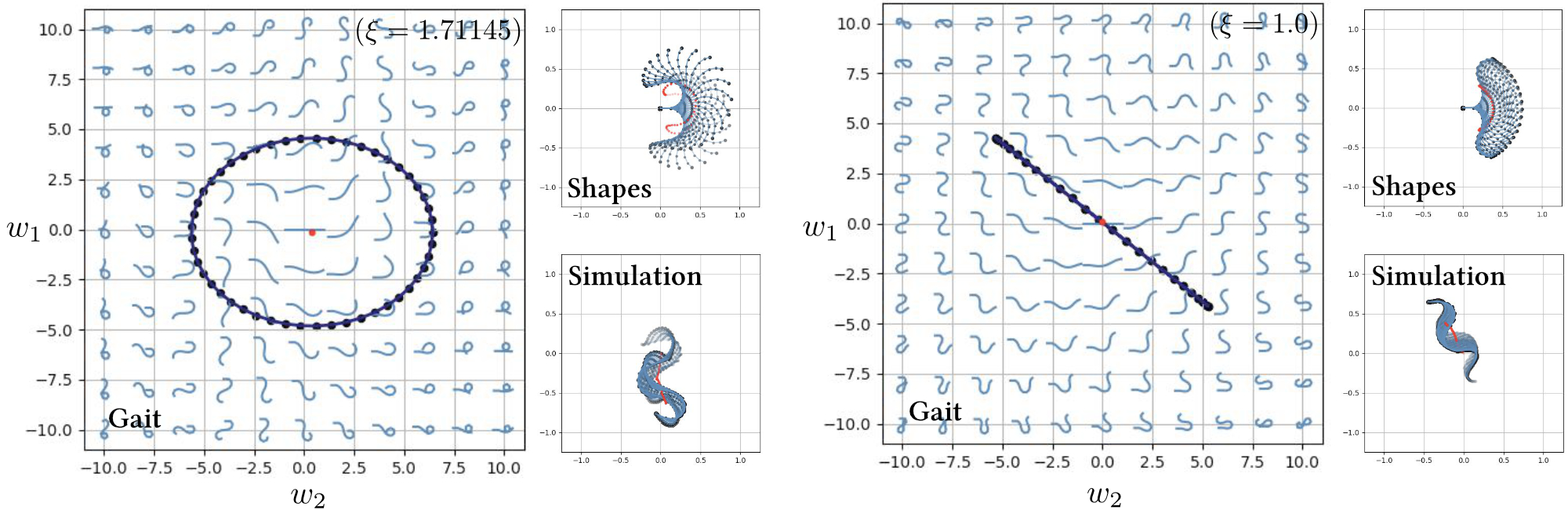}
\vspace{-5mm}
    \caption{The ellipses corresponding to two prototypical gaits within the shape space, accompanied by the shapes of the sequence and motion trajectory.}
       \vspace{-5mm}
\label{fig:GaitEllipses}
\end{figure}

Serpenoid curves are composed from a superposition of a standing and a traveling wave. Typically, planar 
curves are described in terms of their curvature function 
\begin{equation}
     \label{eq:CurvatureInSerpenoidShapeSpace}
     \kappa(s,t) =  w_1(t) \sin(2\pi\xi s) + w_2(t)\cos(2\pi\xi s),
\end{equation}
which, by the fundamental theorem of plane curves~\cite{Pinkall:2024:DG}, uniquely determines the curves' shape up to a rigid body transformation. Here, \(w_1(t)\) and  \(w_2(t)\) are time-dependent coefficients that correspond to the coordinates in the shape space, and \(\xi\) is the spatial frequency of body undulation (\figref{fig:GaitEllipses}). Therefore, the 
serpenoid shape space 
\(\Serp\) is two-dimensional and can be identified with \(\RR^2\) (Figs. \ref{fig:HeroShot}, \ref{fig:GaitEllipses}).

Notably, in contrast to other common choices of shape spaces---such as the space of actuation angles---the dimension of the shape space is independent of the number of actuators on the robot. This is favorable since it lowers the complexity for eventual gait optimization problems on the shape space.

\subsection{Gaits}
\label{sec:Gaits}
\citet{Rieser:2024:GPp} find that the gaits of various undulating living systems tend to follow approximately circular, closed-loop trajectories. Therefore, for the gait design we restrict ourselves to a low-dimensional representation and consider gaits
\begin{equation*}
    [0,1]\ni t \mapsto \begin{pmatrix} w_1 (t) \\ w_2 (t) \end{pmatrix} \coloneqq
    R_\theta\left(a
    \begin{pmatrix}
        \cos(2\pi t) \\
        \sigma\sin(2\pi t)
    \end{pmatrix}\!\right) + \begin{pmatrix} x_{\rm c} \\  y_{\rm c} \end{pmatrix}
\end{equation*}
determined by ellipses embedded in \(\Serp\) (\figref{fig:GaitEllipses}). Here, \(R_\theta\) denotes a rotation in \(\RR^2\) by the angle \(\theta\), \((x_{\rm c}, y_{\rm c})^T\) the center of the ellipse, \(a\) the length of the larger principal axis and \(\sigma\in[0,1]\) the ellipse's flatness. Moreover, we consider the spatial frequency \(\xi\) as a variable. In practice, we discretize those gait ellipses as polygonal curves with their vertices corresponding to time steps (\figref{fig:GaitEllipses}). 

For the purposes of this paper, we consider two kinds of gaits: We either draw uniform samples from the \(6\)-dimensional parameter domain within ``reasonable'' bounds, or we seek for more ``optimal'' gaits. For the latter, we minimize a naive loss function
\begin{equation*}
(\sigma, x_{\rm c}, y_{\rm c}, \theta, a, \xi)\mapsto -\Delta_\mathrm{CoM},
\end{equation*}
where \(\Delta_\mathrm{CoM}\) denotes the magnitude of the translational part of the net displacement of the center of mass resulting from a forward simulation of one full gait cycle with \algref{alg:IntegrateMotion}. A prototypical gait pair is shown before and after optimization for an experiment, simulation, and the elliptical gait profile (SI Movie 0:31-0:44).

We investigate the accuracy of the framework underlying  \algref{alg:IntegrateMotion} 
to describe the motion trajectory of our robotic locomotor in three different ways. First, we qualitatively compare the resulting motion trajectories (\secref{sec:Evaluation}, \figref{fig:TrajectoryComparison}). Second, we examine how accurately performance differences are reproduced (\secref{sec:MappingPerformanceDifferences}, \figref{fig:RatioHeatMaps}). Last, we investigate the possibility of regularizing objective functions to optimize, \eg, the power consumption of gaits (\secref{sec:Regularization}, \figref{fig:Power}). 

\section{Experimental protocol}
\subsection{Snake robot}
\begin{figure}[t]
    \centering
    \vspace{-3mm}
    \includegraphics[width=\linewidth]{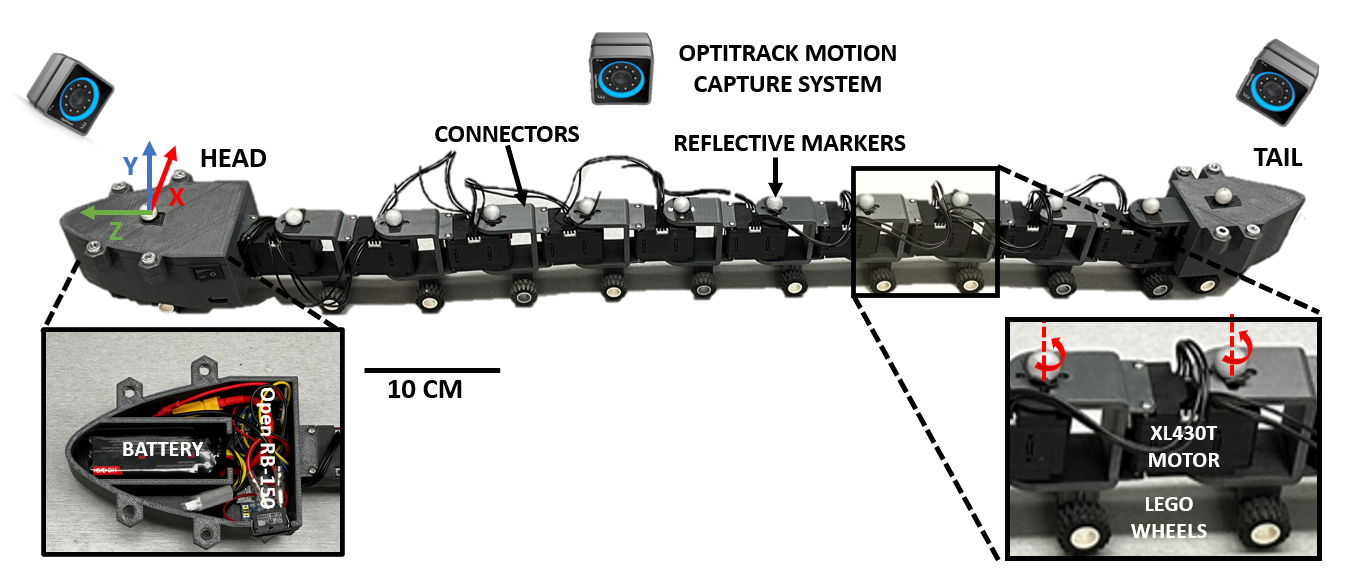}
        \vspace{-4mm}
    \caption{Description of our snake robot which includes ten planar actuators (Dynamixel XL430-W250-T) and rubber wheels that introduce frictional anisotropy. The robot also contains reflective markers at the rotational axis of the motors (marked in red) as well as on the head and tail of the robot which allows each segment to be tracked in real time using the Optitrack Motion Capture system.}
        \vspace{-2mm}
    \label{fig:robot_figure}
\end{figure}

For the experimental setup,
we developed a ten-degree-of-freedom planar snake robot (\figref{fig:robot_figure}). The robot (\(\text{length} = 0.92\,\text{m}\), \(\text{mass} = 1.38\,\text{kg}\)) consists of a chain of Dynamixel XL430-W250-T servo motors connected by 3D-printed ABS plastic connectors printed on a Stratasys F170 3D printer, a Robotis Open-RB 150 control board, and two \(11.1\,\)V \(1500\,\)mAh 120C Drone batteries attached to the head and tail. This type of robot design is well-studied and intentionally simple to avoid introducing unknown sources of error associated with more complex designs.
To account for the frictional anisotropy of the scales, which allow snakes to slither, we attached a pair of passive rubber Lego wheels (\(\text{diameter}= 56\,\text{mm}\), \(\text{thickness}= 11.9\,\text{mm}\), connected by an axle) to each of the connector parts linking the motors (\(18.23\,\)mm from the servo's rotational axis)~\cite{Hirose2009}. Additional wheels were placed beneath the center of mass of both the head and tail of the system (a total of 12 wheel pairs). Functionally, displacements of the wheels in the directions perpendicular to the rolling direction of the wheels generate significantly higher friction than in the tangential directions along the rolling direction.

To investigate whether the underlying geometric model can also be employed to enhance the robot's gait efficiency, we have integrated a power sensor (Adafruit INA260) into the robot. By connecting it in series with the onboard battery, the sensor logs power consumption during operation, enabling us to calculate the Cost of Transport and quantitatively assess efficiency (see \secref{sec:Regularization}).

\subsection{Experimental setup}
We utilized an Optitrack Motion Capture System comprised of eight PrimeX-13 cameras (\(120\,\nicefrac{\rm frames}{\rm sec}\) sample rate) to track the trajectory of the reflective markers that are attached to the robot at the midpoint of the head and tail segments and the axis of rotation of each of the servo motors. Experiments are conducted in a \(3\,\text{m}\times 3\,\text{m}\) arena covered with foam mats. 

\begin{figure}[h]
    \centering
    \includegraphics[width=\columnwidth]{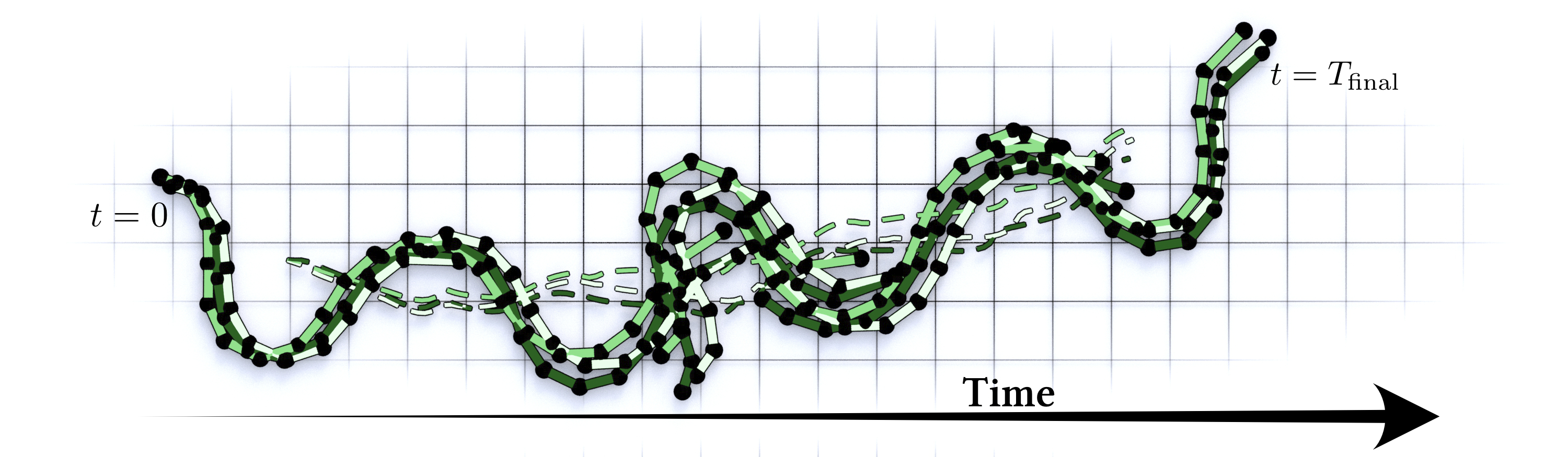}
    \vspace{-7mm}
    \caption{Overlay of three lab trials given the same input data which correspond to the experiment in the center of \figref{fig:TrajectoryComparison}.}
        \vspace{-3mm}
    \label{fig:VarianceFig}
\end{figure}

At the start of each experiment, the robot was placed in the arena with its motors configured to initiate the first configuration of its gait. The robot is equipped with a power located on its head which is used to initiate trials. The simulated gait cycle was discretized and input into the system as a \(50\times10\) matrix. Each gait cycle was executed iteratively three times, with a \(2\)-second pause between cycles to distinguish the multiple cycles within a single trial. We conducted three experiments for each gait cycle and calculated the \emph{mean} and \emph{standard deviation} (STD) of the results to assess the variability and performance consistency across trials (\figref{fig:VarianceFig}).

\subsection{Fitting of material parameters}
\begin{figure}[h]
    \centering
    \includegraphics[width=\linewidth]{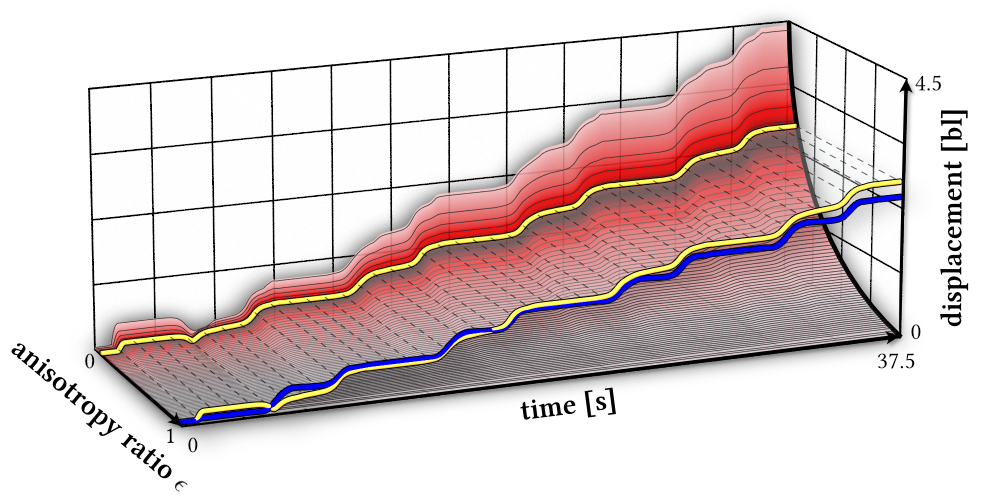}
    \caption{Displacement (body lengths) vs. time graphs for various anisotropy ratios, \(\epsilon\). Red surface shading corresponds with smaller \emph{root--mean--square} (RMS) error between the simulated and lab data (blue). The yellow-highlighted plot shows the simulation using the best--fit \(\epsilon\).}
        \vspace{-5mm}
    \label{fig:EpsFinding}
\end{figure}

Traditional RFT states that motion is predicated on velocity and tangential drag on individual body elements \cite{Gray:1955:PSU, Zhang:2014:TEO}. Our simplified robotic segments can be modeled as point-like contacts on a chosen surface. That is, the tracking points and hence centers--of--mass are at the joints and pivot over the wheelbase (\figref{fig:LocalDissipationTensor}).  We can then use actual weight measurements of the robotic components to determine \(w_j\) (\teqref{eq:LocalDissipationTensor}) and fit the drag--like \emph{anisotropy ratio} \(\epsilon\) as a proxy for tangential drag. The coefficient is inversely proportional to displacement as seen in \figref{fig:EpsFinding}, meaning we can employ, \eg, a binary search~\cite{Cormen:2009:ITA} to find the best--RMS fit between the simulated and laboratory data. For our robot, the average of the fitted values was \(\epsilon_{\rm avg}=0.1865\), which we used for all the experiments presented. 

\section{Evaluation}
\label{sec:Evaluation}
To frame how close our highly simplified model is to reality, we generated twelve random gaits and subsequently optimized them according to \secref{sec:Gaits}. We refer to this set of 24 gaits as the \emph{Simulation} (Sim) data. 

Additionally, when simulating the Sim-data set in laboratory experiments, we collect the corresponding \emph{Experimental} (Exp) data with the help of the motion capture system. A third dataset is obtained by \emph{resimulating} (Resim) the data of the experimental data set. This is because, at each time step, the true angles of the gait were passed to the system as its goal configuration. However, the motors did not always reach their exact goal configuration. We must therefore consider the slightly different shape sequences that have been realized in practice.

 We remark that there is a bijection between the corresponding gaits of each dataset---the simulated gait produces the experimental, which in turn produces the resimulated data.

\begin{figure}[h]
    \centering
\includegraphics[width=\columnwidth]{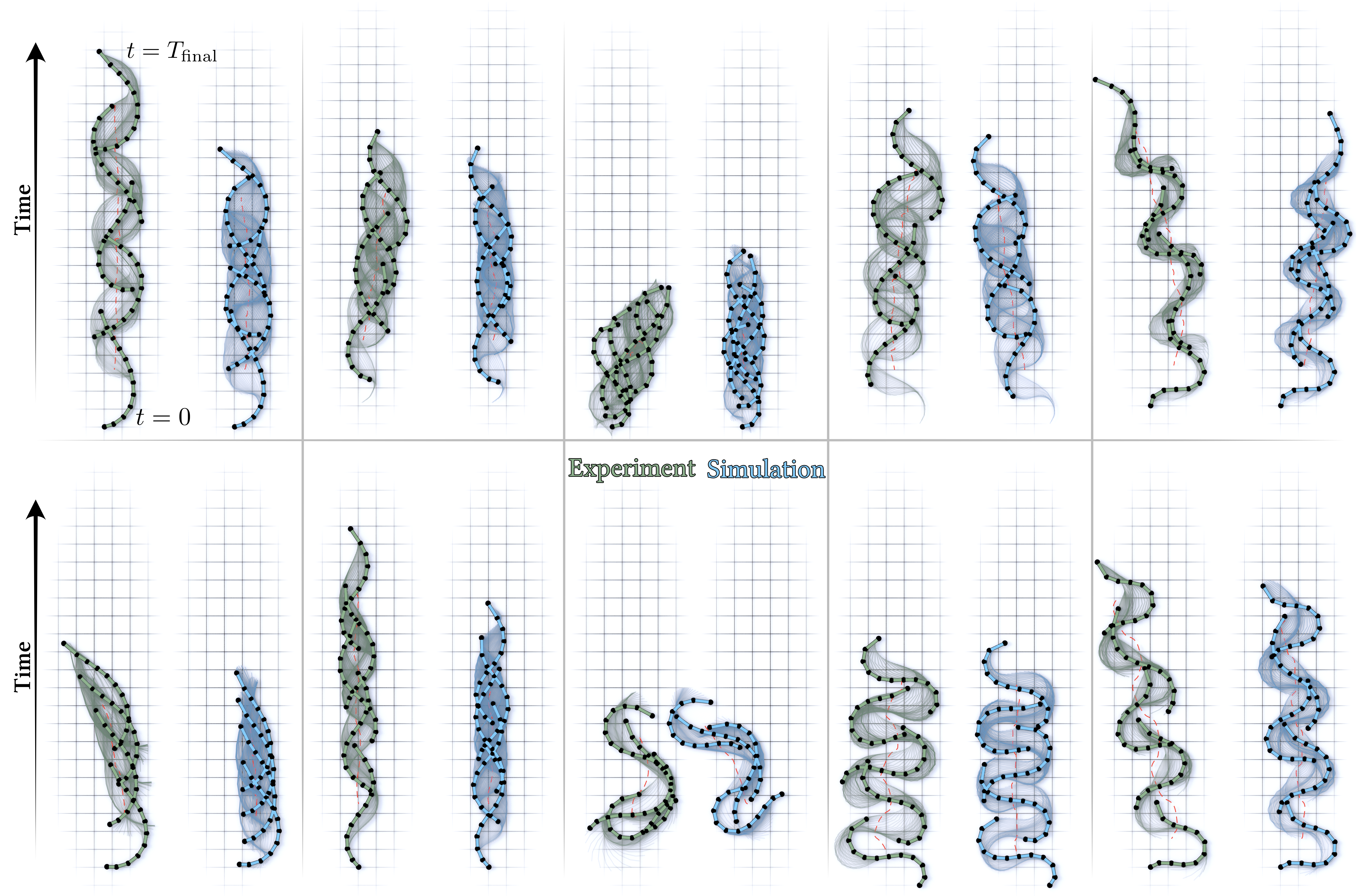}
\vspace{-0mm}
    \caption{Motion trajectories of ten laboratory experiments are shown in green, with simulations corresponding to motion-captured data in blue. The dashed red lines indicate the respective center of mass trajectories.}
        \vspace{-5mm}
\label{fig:TrajectoryComparison}
\end{figure}

\figref{fig:TrajectoryComparison} shows that our highly simplified model fails to reliably reproduce individual trials. Nonetheless, it effectively captures the overall nature of the motion trajectories observed during our laboratory experiments. This is further demonstrated in the SI Movie where all three data classes are shown visually for the gait that achieves maximum CoM displacement (SI Movie 0:03-0:17).


\subsection{Mapping performance differences}
\label{sec:MappingPerformanceDifferences}
To examine the Sim2Real correspondence of the proposed framework, we analyze how accurately differences in the performances of gaits are mapped.  In theory, gaits that excel or underperform in simulations should exhibit similar performance trends in real-world experiments. We compare each of the gaits---Sim, Exp and Resim---to every other gait in the same class, and consider the \emph{performance ratios} \[\delta((i,j),X)\coloneqq\left[\tfrac{\Delta_\text{CoM}(\text{gait}_i)}{\Delta_\text{CoM}(\text{gait}_j)}\right]_{i,j\in\mathrm{class}\, X}\] of their net displacements, which is a unitless metric correlating the gait performances within each class. For the experiments, each gait was tested through nine trials and the average CoM displacement is considered the experimental ground truth.


\begin{figure*}
    \centering
    \includegraphics[width=\textwidth]{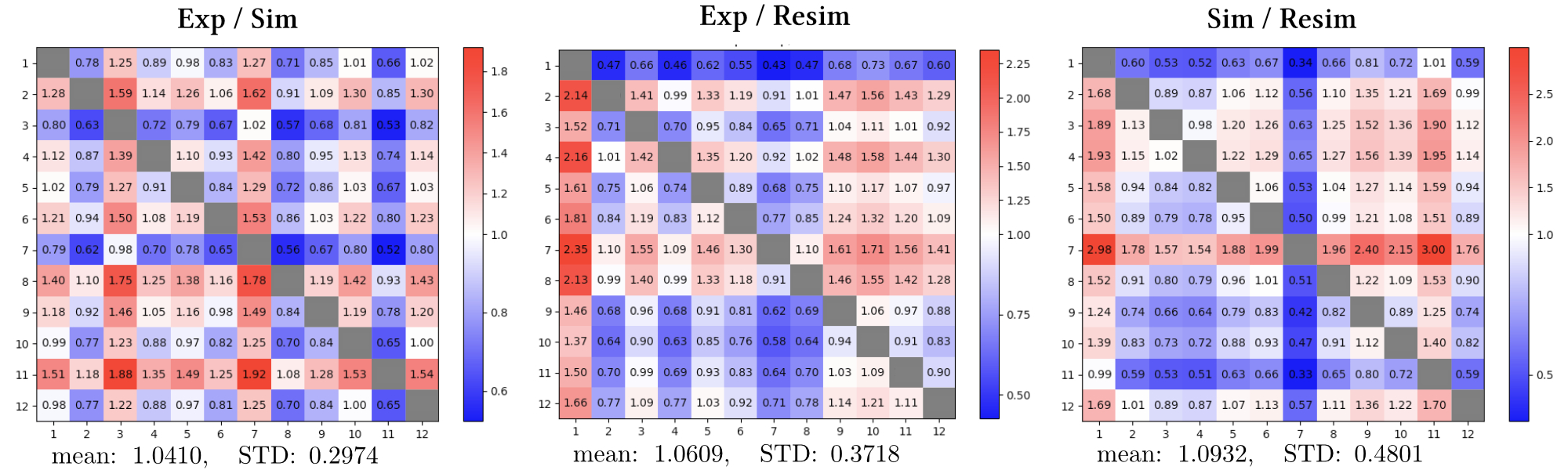}
    \vspace{-5mm}
    \caption{These heat maps show the pairwise quotients of the gait performance ratios across the Experimental (Exp), Simulated (Sim), and Resimulated (Resim) classes. The statistical summary below the maps provides the mean value and standard deviation of cross-class comparisons, focusing on the consistency between experimental data and simulations. }
    \label{fig:RatioHeatMaps}
\end{figure*}

Then, for corresponding gait pairings \((i,j)\) of two classes \(X\) and \(Y\), we consider the quotients \[\Xi((i,j),X,Y)\coloneqq\tfrac{\delta((i,j),X)}{\delta((i,j),Y)}\] of the performance ratios to determine how well differences in gaits were mapped from one class to another. This quotient naturally captures all disparities in the performance ratios caused by the class transitions. The data is displayed in a Heat Map in~\figref{fig:RatioHeatMaps}, with the mean and the STD being computed without the data of the major diagonal where gaits are compared against themselves.

The comparison between experimental, simulated, and resimulated gaits provides insight into the effectiveness of our simulation framework. The Exp/Sim comparison shows strong alignment (\(\text{mean} = 1.0410\), \(\text{STD} = 0.2974\)), indicating that our initial simulations closely follow experimental data. However, the Exp/Resim comparison reveals slightly more variability (\(\text{mean} = 1.0609\), \(\text{STD} = 0.3718\)), suggesting that resimulating based on tracked shapes introduces some divergence. Finally, the Sim/Resim comparison (\(\text{mean} = 1.0932\), \(\text{STD} = 0.4801\)) suggests that while the two simulated datasets share general trends. The resimulation leads to a large variability, which is another indication that the robot does not achieve the desired shapes accurately.

Since the STD of all the between-class comparisons is significant, it is unlikely that individual trials will be accurately reproduced. But since the average of all trials was approximately \(1\), we conclude that \algref{alg:IntegrateMotion} is effective at mapping displacement from simulation to reality in the aggregate, if the amount of trials is sufficiently large. 


\subsection{Regularization}
\label{sec:Regularization}

Another consideration is power efficiency. In robotic systems, battery power is at a premium and we tested the ability to improve gait efficiency. 

Experimentally we know that energy is dissipated into the environment through the frictional resistance of the wheels. Therefore, we penalize the total energy dissipation (see \teqref{eq:VariationalEnergy}) in the simulation by introducing a penalty term to the objective of the optimization. That is, we minimize \[-\Delta_\mathrm{CoM} + c\,\cE,\] 
where the \emph{Dissipation Coefficient} \(c>0\) is a positive constant. This punishes energy loss while still attempting to maximize the CoM displacement. 
Additionally, we introduce the possibility for a constraint on the spatial frequency to generate gaits with a prescribed wavelength.

\begin{figure}[t!]
    \centering
\includegraphics[width=\columnwidth]{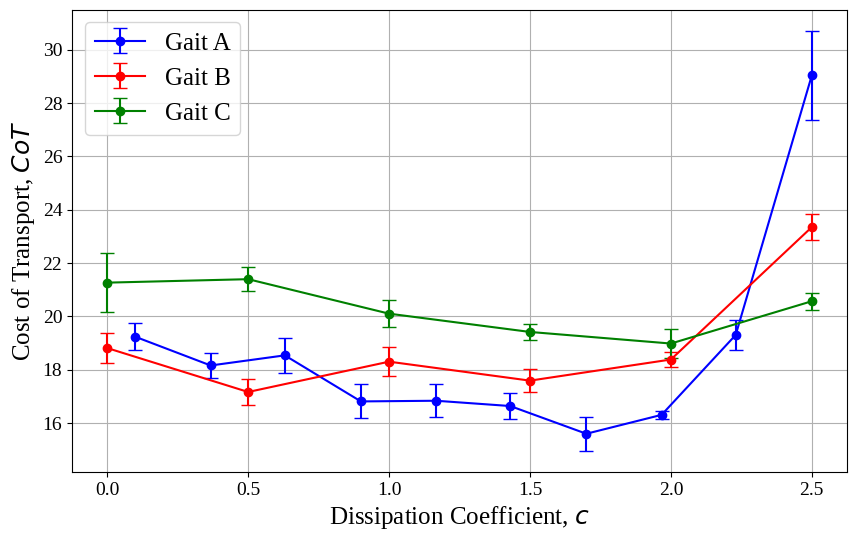}
    \vspace{-7mm}
    \caption{The effectiveness of penalizing dissipation in the experiment for three representative gaits (A, B, C). For each gait, the mean Cost of
    Transport is a function of the Dissipation Coefficient $c$, and the standard deviation of the Cost of Transport is represented by error bars. }
    \label{fig:Power}
\end{figure}

In a robotic system, a key metric for evaluating gait efficiency is the \emph{Cost of Transport}  \[\mathrm{CoT} \coloneqq\nicefrac{P}{m g v},\] where \(P\) denotes the \emph{power} \([\nicefrac{\rm J}{\rm s}]\), \(m\) is the \emph{mass} \([{\rm kg}]\), \(g\) is \emph{gravitational acceleration} \([\nicefrac{\rm  m}{{\rm s}^2}]\) and \(v\) is the \emph{velocity} \([\nicefrac{\rm m}{\rm s}]\). To minimize the cost of transport, a gait needs to balance minimizing power consumption while achieving significant displacement. To measure this, we ran three trials with three gaits each and measured average power consumption. Simultaneously, the CoM displacement was recorded using the motion capture system. The data for each point displayed in \figref{fig:Power} is a summary of information of 9 power and displacement trials.

To determine if the optimization with an additional penalty term leads to
efficiency improvements, 
we took a range of values for $c$, from \(0\) to \(2.5\), for three representative gaits. For Gaits A and B, only the dissipation penalty is introduced to the optimization increase in weight. While for Gait C, a spatial frequency constraint 
was added. The experimental results indicate that the inclusion of penalty terms can improve the robot's performance in terms of CoT. 

When $c$ is low, the gaits obtained from optimization will barely change. 
On the other extreme, when the penalty on dissipation is increased beyond a threshold, the optimal actuation of the robot becomes very subtle, and thus the CoM displacement drops significantly, increasing the CoT. Additionally, as the motor movements become more subtle, static friction is encountered at a higher rate---increasing the required torque power consumption. However, when $c$ is between \(1.0-2.0\), a balance is struck between power consumption and displacement. The strongest evidence of this is in Gait A at $c=1.7$ where we see a 23\% improvement in the CoT as displayed (SI Movie 0:17-0:31). We see then, that penalizing dissipation is reflected in simulated results.

\subsection{Limitations and Challenges}
There are two major limitations of the proposed approach in its current form: Slipping and reliance on motion capture data. Slipping typically occurs when the passive rubber wheels fail to maintain consistent traction with the surface, leading to deviations in the robot's trajectory compared to the predicted one. This issue is primarily due to frictional inconsistencies on the experimental surface, as well as motor control limitations that prevent them from precisely reaching desired states. Slipping can introduce compounding errors in CoM displacement measurements, making it difficult to accurately resimulate the experimental trials. To mitigate this, future work could explore refining contact surface properties for consistent traction or upgrading the wheel design to something like a kirigami skin \cite{rafsanjani2018kirigami,Branyan2020}. Additionally, feedback control systems that dynamically adjust the robot's movement in response to slipping as well as experimentation on various surfaces would provide valuable insights into gait efficiency and accuracy.

We note that the relatively high anisotropy ratio fitted to our experimental values betrays this slipping. That is, with a wheeled system that reinforces tangential motion, we expect a low (\(\epsilon \approx 0\)) value. The lack of normal friction is then reflected in the simulation framework as higher tangential dissipation and can be effectively seen as a metric for the slipping frequency. 


Moreover, our approach currently relies on external motion capture measurements for calibration. While our gait generation and improvement pipeline is powerful because it relies on only one physical parameter, $\epsilon$, this could be problematic in dynamic environments. We conducted experimental trials using a motion capture system to analyze the behavior of the current environment and evaluate its dissipative properties. Ideally, real-time environmental feedback would enable the system to compute the anisotropy ratio onboard and dynamically adjust its gait in response to changing conditions.


\section{Conclusion and Outlook}
We successfully demonstrated the implementation of a variational integrator based on geometric mechanics to model the undulatory motion of a snake-like robot. By characterizing the dissipative losses to the environment, the proposed method effectively characterizes the motion resulting from a sequence of robot configurations. It greatly simplifies the computational cost of modeling complex robot dynamics, particularly when compared to current physics engines \cite{collins2021review}. We validated the accuracy of the simulation through the mean comparison of various gaits against experimental data and demonstrated strong correspondence of CoM displacement. Additionally, we demonstrated the ability to regularize gaits and improve efficiency as measured by the Cost of Transport by up to \(23.2\%\). 

Although our results indicate an effective ability to qualitatively reproduce real-world results, there are many ways in which the gait generation pipeline could be improved. In its current state, \algref{alg:IntegrateMotion} can be used to evaluate the effectiveness of a gait through a full forward simulation. This is only tractable due to the algorithm's computational efficiency. For more reliable gait optimizations, more sophisticated optimization methods relying on, \eg, back-propagation should be explored. 
Nonetheless, we envision that the gaits we generate could be used as training data to convert a sequence of motor positions directly to a sequence of CoM displacement.  

Neural network architectures such as Recurrent Neural Networks (RNNs) \cite{Cho:2014:EMNLP}, Long Short-Term Memory networks (LSTMs)\cite{Hochreiter:1997:NC}, and Transformers \cite{Vaswani:2017:CoRR} have proven highly effective in sequence-to-sequence optimization tasks, making them well-suited for mapping motor position sequences to CoM displacement in robotic systems, enabling efficient and accurate gait prediction without the need for full forward simulations. In this way, we could infer the neural network to detect candidate gaits and evaluate performance before a robot is built. The model may not replicate every trial precisely; however, the experimental behavior can be accurately captured through a large sample size, as discussed in \secref{sec:MappingPerformanceDifferences}.

This research opens new avenues for efficient and adaptable robotic locomotion, with potential applications in reinforcement learning, inverse design, and real-world robotic deployment across diverse environments.


%



\section*{Acknowledgment}
We would like to thank the Naughton Undergraduate Fellowship program for supporting Cora Keogh. Additional support was provided by SideFX software.




\bibliographystyle{IEEEtranN} 

\end{document}